\documentclass[journal]{IEEEtran}

\IEEEoverridecommandlockouts
\usepackage{cite}
\usepackage{amsmath,amssymb,amsfonts}
\usepackage{algorithmic}
\usepackage{graphicx}
\usepackage{textcomp}
\usepackage{xcolor}
\usepackage{subcaption}
\usepackage[colorlinks=false,hidelinks,bookmarks=false]{hyperref}
\usepackage[letterpaper, top=0.75in, bottom=0.75in, left=0.625in, right=0.625in]{geometry}
\usepackage{array}
\usepackage{soul}  
\usepackage{times}
\usepackage{booktabs}
\usepackage{multirow}
\usepackage{caption}
\usepackage{subcaption}
\usepackage{setspace}
\usepackage{algorithm}
\usepackage{float}
\usepackage{microtype}
\usepackage{enumitem}
\usepackage{tabularx}
\usepackage{boldline}

\def\BibTeX{{\rm B\kern-.05em{\sc i\kern-.025em b}\kern-.08em
    T\kern-.1667em\lower.7ex\hbox{E}\kern-.125emX}}

\begin{document}

\title{A Two-Stage Time-Aware Transformer for Short-Horizon AECOPD Risk Prediction
 }

\author{Dongyang Wang, Weihao Qu, Ling Zheng, and Haowen Pan
\thanks{D. Wang, W. Qu, and L. Zheng are with the Department of Computer Science and Software Engineering, Monmouth University, West Long Branch, NJ, USA. E-mail: \{s1382037, wqu, lzheng\}@monmouth.edu}
\thanks{H. Pan is with Changzhou Yaoyuanxing Electronic Technology Co., Ltd., China. E-mail: hpan2414@gmail.com}}

\newcommand{\wq}[1]{\textcolor{black}{#1}}

\maketitle

Acute exacerbation of chronic obstructive pulmonary disease (AECOPD) can worsen rapidly, making timely prediction a clinical priority. Most existing machine learning approaches rely on episodically collected clinical variables, introducing delays that limit their practical utility in home monitoring settings. Home ventilators offer a lower-latency alternative, producing a near-continuous record of respiratory status during daily use. However existing ventilator-based approaches either compress the waveform into handcrafted features or focus primarily on binary risk classification, leaving the timing of an impending event unresolved. In this paper, we present a two-stage framework that operates directly on raw pressure and flow waveforms from the most recent seven days of home ventilator use. The first-stage classification model identifies patients at high risk of a severe exacerbation. The second-stage regression model then estimates how many days remain before the event occurs. Our experimental results demonstrate that the two-stage model outperforms traditional baseline models on both risk classification and time-to-event estimation, with our selected Stage~1 classifier achieving F1 = 0.91 and our Stage~2 regression model achieving RMSE = 1.00 days and $R^2$ = 0.76, giving clinicians both an early warning and actionable lead time before a severe exacerbation occurs.

\section{Introduction}
\label{sec:intro}

Acute exacerbation of chronic obstructive pulmonary disease (AECOPD) refers to
a sudden worsening of respiratory symptoms beyond normal day-to-day
variation \cite{rueda2024machine}. AECOPD substantially worsens
quality of life and is associated with increased hospitalization and
mortality \cite{lenoir2023mortality}. We therefore view early detection as a
clinical priority, but most existing machine learning models rely on
clinical and laboratory inputs such as electronic health records,
spirometry, blood gas analyses, and symptom questionnaires
\cite{kor2022explainable}. Because these signals are collected
episodically, they introduce a delay between physiological
deterioration and risk detection that is especially problematic for a
condition that can worsen rapidly.

Home ventilator waveforms provide a lower-latency alternative. Patients
with severe COPD often use home non-invasive ventilators for several
hours per day, and the resulting \texttt{pressure} and \texttt{flow}
signals offer a near-continuous record of respiratory status in the home
environment. Unlike clinic-derived measurements, these waveforms are
available during routine daily use and can therefore support short-horizon monitoring without waiting for the next hospital visit or test.

Recent machine learning studies have shown that respiratory and clinical
time-series data carry predictive value, including explainable COPD risk
models \cite{kor2022explainable}, data-driven COPD flare-up detection
\cite{rueda2024machine}, and modern transformer models for multivariate
time-series representation learning \cite{foumani2024tape,ICLR2025_2b187165}.
However, an important gap remains: many existing AECOPD approaches still
aggregate the waveform into handcrafted summary features or otherwise
weaken the temporal structure of the raw signal, making it harder to
capture the short-horizon dynamics that precede severe exacerbation.

In this paper we address two clinical questions. First, given a rolling window of home ventilator recordings, can a model identify patients at high risk of an imminent severe AECOPD event? Second, for patients
already identified as high risk, how many days remain before that
severe exacerbation event? We view the first
question as supporting early warning, whereas the second provides actionable
lead time for intervention.

To answer these questions, we use a two-stage model inspired by recent
transformer-based time-series modeling work with temporal representation
learning \cite{foumani2024tape,ICLR2025_2b187165}.
At a high level, the approach operates directly on raw \texttt{pressure}
and \texttt{flow} waveforms over a 7-day window, first producing a
binary high-risk decision and then estimating time to event only for the
patients identified as high risk. We design the pipeline to preserve the temporal structure of the raw waveform signal and to mirror the natural clinical workflow: screen for risk first, then estimate urgency.

Our main contributions are as follows:
\begin{enumerate}[leftmargin=*, label=(\roman*)]
    \item Instead of extracting all ventilator variables into handcrafted features, we
          keep the two primary waveform channels, \texttt{pressure} and
          \texttt{flow}, over a 7-day window. These channels directly reflect
          patients' respiratory mechanics, triggering, cycling, and air
          trapping. This method preserves the
          breath-level temporal dynamics, which are indispensable for short-horizon deterioration
          prediction.

    \item Time-Aware Transformer encoder adapted for
          raw respiratory time-series data, learning patient representations
          from \texttt{pressure} and \texttt{flow} waveforms that can be
          reused by downstream classifiers and regression models.

    \item A two-stage prediction pipeline in which
          Stage~1 combines the learned representations with downstream classifiers for
          high-risk classification, and Stage~2 applies a separately trained
          time-aware regression model for time-to-event estimation in
          patients identified as high risk.

    \item An empirical evaluation in which our selected
          32-dimensional time-aware Stage~1 configuration uses logistic regression as the primary classifier (F1 = 0.91 for label~1) and XGBoost as a secondary classifier for stability verification, while our selected 64-dimensional Stage~2 model achieves RMSE = 1.00 days, MAE = 0.87 days, and $R^2$ = 0.76
          on the held-out test set.

\end{enumerate}

The remainder of the paper is structured as follows.
Section~\ref{sec:data} describes the dataset and preprocessing pipeline.
Section~\ref{sec:method} presents the architecture of the two-stage model.
Section~\ref{sec:experiments} reports experimental results.
Section~\ref{sec:discussion} discusses clinical implications, limitations,
and future directions.

\section{Related Work}
\label{sec:related}

Prior work on AECOPD prediction has largely relied on structured
clinical variables, telemonitoring summaries, or hand-crafted features
rather than raw ventilator waveforms. Recent COPD-focused studies have
used explainable clinical risk models, data-driven flare-up detection,
and day-to-day home noninvasive ventilation parameters to predict or
characterize exacerbation risk
\cite{kor2022explainable,rueda2024machine,jiang2021nppv,wu2021upcoming7days}.
These studies support the feasibility of early AECOPD prediction, but
they generally depend on engineered features, intermittent measurements,
or multimodal summaries that may still introduce latency or discard part
of the original temporal waveform structure. In addition, most AECOPD
prediction studies formulate the task mainly as binary classification,
which can indicate whether risk is elevated but not how soon an event is
likely to occur.

More recently, transformer-based time-series models have advanced rapidly,
including improved positional encoding for multivariate time-series
classification, general-purpose time-series representation learning, and
new architectures for long-horizon temporal modeling
\cite{foumani2024tape,ICLR2025_2b187165}. Recent studies have also emphasized scalable explainable AI
and behavior-guided learning in intelligent systems~\cite{tyrovolas2026efficient,li2026behavior,qu2026multimodal}.

At the same time, prior long-horizon forecasting work has questioned
whether transformer architectures consistently outperform simpler linear
baselines for time-series forecasting \cite{zeng2023transformers}.
A related jump-point time-aware transformer for AECOPD prediction compressed
the respiratory waveform into sparse event representations before
classification \cite{jumpbaseline2025}. In contrast, our work keeps the
raw \texttt{pressure} and \texttt{flow} waveforms, uses explicit temporal
encoding over the last seven days, and extends beyond binary
classification to a two-stage pipeline that also estimates time to event.
This design is intended to preserve continuous respiratory deterioration
patterns that are especially important for downstream regression and alert
timing. Prior COPD exacerbation prediction studies have commonly used XGBoost as a strong machine-learning baseline for remote-monitoring and near-future AECOPD prediction~\cite{yin2024inhome,liao2024nearfuture}. Therefore, we include XGBoost as a non-transformer comparator and stability-check model.

\section{Data and Preprocessing}
\label{sec:data}

\subsection{Dataset}

The dataset initially comprised continuous one-month respiratory
time-series recordings from 87 COPD patients collected via
daily-use home non-invasive ventilators between 2023 and 2025 (42
patients from 2023, 10 from 2024, and 35 from 2025). Of these,
57 patients were assigned to label~0 (no acute exacerbation),
whereas 30 patients were assigned to label~1 (severe AECOPD
requiring emergency or intensive-care-level treatment). 2 patients
were excluded before model development because their recordings did not
provide sufficient temporal coverage for the required analysis windows. The
final analysis contained 85 patients. Daily
usage ranged from 4 to 12 hours,
producing between 72,000 and 220,000 rows per day at a sampling rate of
five readings per second. Each row contains eight columns: timestamp,
flow, pressure, peripheral oxygen saturation (SpO$_2$), respiratory rate,
tidal volume, minute ventilation, and system leak. Timestamp records the
exact date and time of the measurement. Flow, minute ventilation, and leak
are recorded in liters per minute. Pressure is recorded as circuit air
pressure in centimeters of water (cmH$_2$O). SpO$_2$ records peripheral oxygen saturation.
Respiratory rate is measured in breaths per minute. Tidal volume is the
exhaled volume per breath cycle in mL.

The cohort was split into 48 training, 15 validation, and 22 test patients. Training and validation sets were stratified to preserve the natural 2:1 negative-to-positive ratio (training: 32 label-0, 16 label-1; validation: 10 label-0, 5 label-1), reflecting true clinical prevalence and avoiding artificial resampling. The test set comprises 17 label-0 and 5 label-1 patients.

These ventilator waveform data were obtained from a collaborating hospital, deidentified before analysis,
and used under institutional ethics approval and consent procedures to be confirmed in the final manuscript.

Code is available at: \url{https://github.com/WangPage81/AECOPD-home-ventilator-prediction/tree/main}. Patient data cannot be shared due to privacy restrictions.

\subsection{Preprocessing}

We use a targeted preprocessing strategy that retains only the \texttt{pressure} and \texttt{flow} columns from the last 7 days before the prediction reference point.

Rather than using all seven physiological ventilator variables, we
retain only the two most clinically representative input channels:
\texttt{pressure} and \texttt{flow}. No handcrafted feature extraction
or waveform compression is applied. We selected these two channels because they are the primary ventilator
scalar waveforms routinely used to assess airway resistance,
respiratory mechanics, patient effort, triggering, cycling, and air
trapping \cite{hess2005waveforms,fernandez2006pressure,hamahata2020flow}.
In addition, flow-derived measures such as peak expiratory flow have
shown clinical value for detecting COPD exacerbation and assessing
hospitalization-level deterioration \cite{cen2019pef,cen2022pef}.

\begin{figure}[t]
    \centering
    \includegraphics[width=0.88\columnwidth]{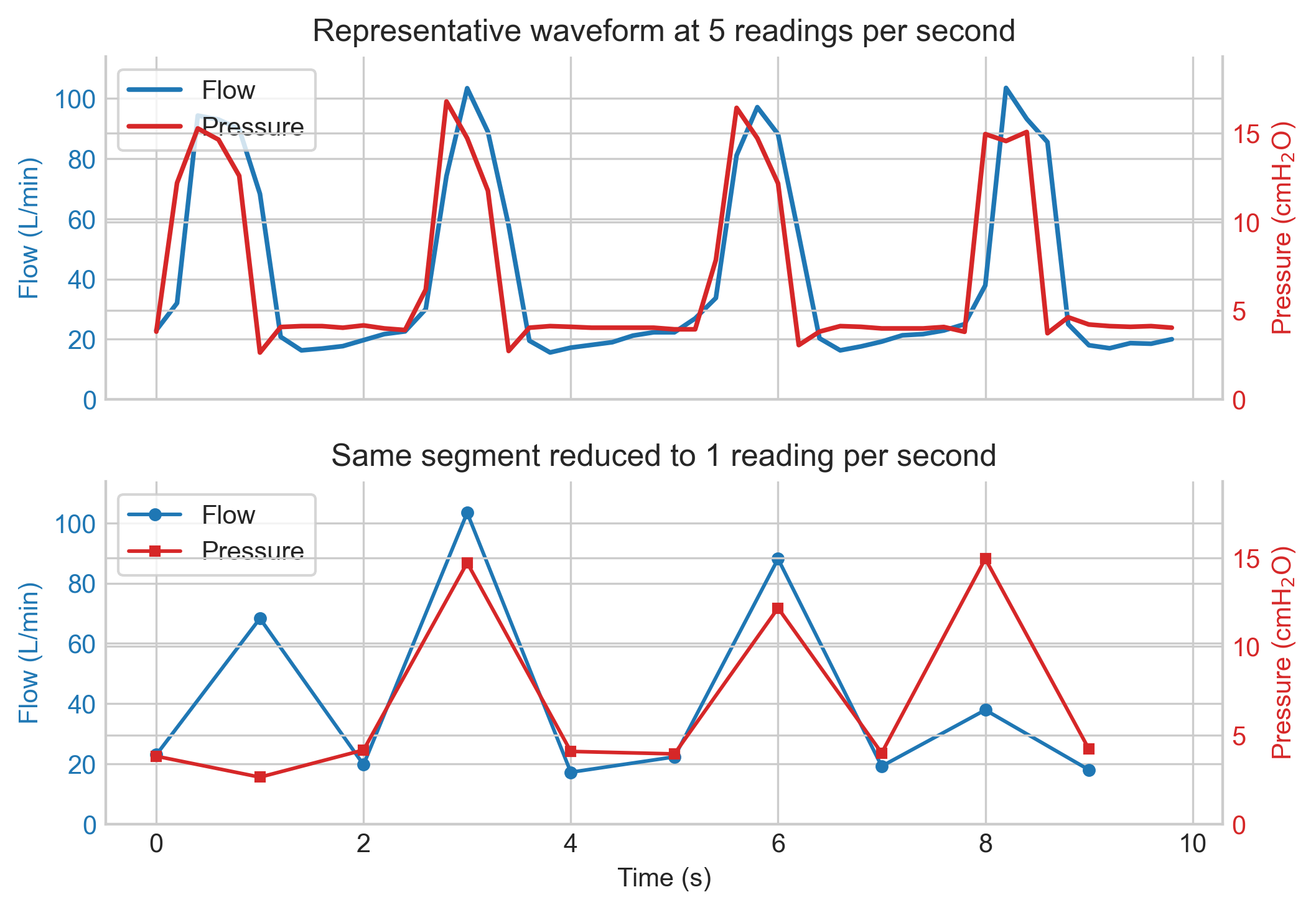}
    \caption{Representative raw ventilator waveform segment comparing
    the original 5~Hz recording with the same segment reduced to 1
    reading per second.}
    \label{fig:waveform_example}
\end{figure}

Figure~\ref{fig:waveform_example} shows a representative short segment
of the raw ventilator waveform used in this study. Because the device
records five entries per second, the data preserve fine-grained
breath-by-breath temporal variation in \texttt{flow} and
\texttt{pressure}, including rapid within-breath changes and cycle-to-cycle
transitions. Figure~\ref{fig:waveform_example} also compares this 5~Hz
segment with the same signal reduced to 1 reading per second, showing
that much of the waveform shape is lost even before any additional
compression is applied. In contrast, jump-point data retain
only timestamps where the absolute difference between the current value
and the previous retained value exceeds a threshold. This sparse encoding cannot fully preserve the continuous
5~Hz waveform structure used by our model.

We further restrict the input to the most recent seven days before the
prediction reference point. This short-horizon window is motivated by
both clinical and practical considerations. Clinically, ventilator-based
changes associated with impending exacerbation can emerge during the
week before hospitalization, and prior studies have explicitly examined
abnormal respiratory patterns within a 7-day pre-AECOPD period and
developed models for predicting AECOPD in the upcoming 7 days
\cite{jiang2021nppv,wu2021upcoming7days}. Practically, using the last
seven days keeps the raw-sequence length manageable for transformer
processing and keeps the model focusing on the period most relevant to
short-horizon deterioration assessment.

\section{Methodology}
\label{sec:method}

\begin{figure*}[t]
    \centering
    \includegraphics[width=0.78\textwidth]{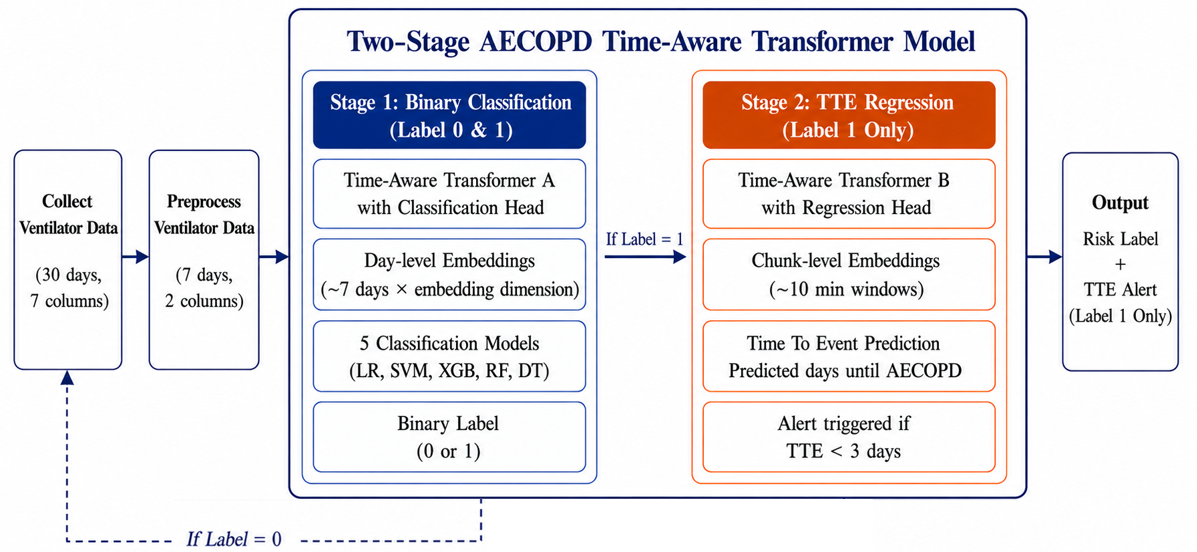}
    \caption{Architecture of the Two-Stage AECOPD Time-Aware Transformer Model.}
    \label{fig:architecture}
\end{figure*}

Figure~\ref{fig:architecture} illustrates the architecture of the
Two-Stage AECOPD Time-Aware Transformer Model. The pipeline first
preprocesses 30-day raw ventilator data by selecting only the last
seven days of \texttt{pressure} and \texttt{flow}. It then applies two
\emph{separate} Time-Aware Transformers: Transformer~A for binary
classification and Transformer~B for time-to-event (TTE) regression.

\subsection{Separate Time-Aware Transformers}
\label{sec:encoder}

Although Transformer~A and Transformer~B share the same Time-Aware
Transformer architecture family, they are trained independently rather
than through transfer learning. This design choice reflects the fact
that the two stages solve different learning problems under different
data distributions. Transformer~A is trained on all 85 patients to
separate high-risk from low-risk ventilator patterns, so its parameters
are shaped by a binary discrimination objective and by exposure to both
label~0 and label~1 examples. Transformer~B, by contrast, is trained
only on the 29 label~1 patients and optimized for a continuous
time-to-event regression target. The countdown target used in Stage~2
(from day~6 to day~0 before exacerbation) has no analogue in Stage~1,
so initialising Transformer~B from Transformer~A would introduce a
classification-oriented inductive bias without a clear task-level
benefit. Keeping the two transformers separate therefore allows each
model to be optimized for its own objective without cross-task
interference.

Let a patient's preprocessed raw sequence be denoted
$\mathcal{S} = \{(e_1, v_1, \tau_1), (e_2, v_2, \tau_2), \ldots, (e_N, v_N, \tau_N)\}$,
where $e_i$ is the channel type, $v_i$ is the scalar sample value, and
$\tau_i$ is the elapsed time from the start of the first recording.
We use elapsed time because prior transformer time-series work has shown
that absolute or elapsed position encoding preserves sequential order.
Time2Vec-style temporal representations can also improve temporal
modeling performance \cite{foumani2024tape,kazemi2019time2vec}.

\paragraph{Input Embeddings.}
Each token is represented as the sum of three embeddings:

\begin{equation}
    \mathbf{x}_i = \mathbf{E}_{\text{type}}(e_i)
                 + \mathbf{E}_{\text{value}}(v_i)
                 + \mathbf{E}_{\text{time}}(\tau_i)
    \label{eq:embedding}
\end{equation}

\noindent where $\mathbf{E}_{\text{type}} \in \mathbb{R}^{|\mathcal{V}| \times d}$
is a learned type embedding matrix, $\mathbf{E}_{\text{value}}$ is a
linear projection from scalar to $d$-dimensional space, and
$\mathbf{E}_{\text{time}}$ is a Time2Vec-style temporal encoding
\cite{kazemi2019time2vec} applied to the elapsed-time index:

\begin{equation}
    [\mathbf{E}_{\text{time}}(\tau_i)]_j =
    \begin{cases}
        w_0 \, \tau_i + b_0 & j = 0 \\
        \sin(w_j \, \tau_i + b_j) & j \geq 1
    \end{cases}
    \label{eq:time2vec}
\end{equation}

Using elapsed time from the start provides a stable, monotonic
temporal reference within every sample window. This is particularly
useful for downstream TTE regression because it aligns within-window
progression across patients without introducing irrelevant absolute
clock times. A learnable positional encoding is added prior to the
transformer encoder layers. Each transformer has $L = 2$ layers of
multi-head self-attention followed by feed-forward networks, with
$H = 4$ attention heads and embedding dimension $d \in \{32, 64, 128\}$.

\subsection{Stage 1: Binary Classification}
\label{sec:classification}

Stage~1 uses Time-Aware Transformer~A with classification head and is trained on all 85
patients. The raw sequence is segmented into
day-level segments over the last seven days. Each day is
encoded independently, and the resulting seven day-level embeddings are
concatenated into a patient feature vector
$\mathbf{f} \in \mathbb{R}^{7d}$. This vector is then passed to five
standard classifiers: Logistic Regression (LR), Support Vector Machine
(SVM), Decision Tree (DT), Random Forest (RF), and XGBoost (XGB), as shown in the left branch of Figure~\ref{fig:architecture}.

Transformer~A is trained end-to-end with binary cross-entropy (BCE)
loss with positive class weighting:

\begin{equation}
    \mathcal{L}_{\text{cls}} =
        -\frac{1}{N} \sum_{i=1}^{N}
        \left[ w_+ \, y_i \log \hat{p}_i
             + (1 - y_i) \log (1 - \hat{p}_i) \right]
    \label{eq:bce}
\end{equation}

where $w_+ = N_0 / N_1$ is the ratio of negative to positive samples.
The output of Stage~1 is a binary risk label.

\subsection{Stage 2: Time-to-Event Regression}
\label{sec:regression}

Stage~2 uses Time-Aware Transformer~B with regression head, which is trained
separately and is not weight-shared with Transformer~A.
This stage uses only the 29 label~1 patients remaining after cohort
filtering. Unlike the classification pipeline, the regression pipeline
does not reuse saved day-level embeddings from Stage~1. Instead,
Transformer~B operates directly on raw preprocessed chunks and predicts
the time to event, defined here as the time to ICU-level exacerbation,
which we refer to here as days until exacerbation (stored in the code as \texttt{countdown\_days\_target}).

Following the right branch of Figure~\ref{fig:architecture}, the
regression branch converts raw sequences into chunk-level embeddings
using approximately 10-minute windows. The regression head then maps
each chunk representation to a scalar TTE target:

\begin{equation}
    \hat{y}_{\text{reg}} = g(\mathbf{z}_{\text{chunk}})
    \label{eq:regression}
\end{equation}

where $g$ is a regression head. Targets are transformed by
$\log(1 + \text{days})$ and min-max scaled to $[0, 1]$. The regression
model is trained with mean squared error (MSE) loss on the scaled
targets:

\begin{equation}
    \mathcal{L}_{\text{reg}} =
        \frac{1}{N} \sum_{i=1}^{N} (\hat{y}_i - y_i)^2
    \label{eq:mse}
\end{equation}

At inference, the inverse transforms recover predicted days until
AECOPD. The deployment logic follows the two-stage workflow shown in
Figure~\ref{fig:architecture}: each rolling seven-day window is first
processed by Stage~1. If Stage~1 predicts label~1, the patient is passed
to Stage~2, which outputs the TTE estimate; an alert is triggered when
the predicted TTE is less than 3 days. If Stage~1 predicts label~0, no
TTE regression is performed for that window, and the system continues
monitoring the patient with the next rolling seven-day window.

\subsection{Training Details}

Transformer~A and Transformer~B use the same Time-Aware Transformer
architecture family but are trained as two fully separate models with
different data splits, objectives, and checkpoints. Transformer~A is
trained for classification on all 85 patients, whereas Transformer~B is
trained only on the 29 label~1 patients used for TTE regression. For
example, the 32-dimensional transformer uses 4 attention
heads, a feed-forward hidden dimension of 128, and
2 encoder layers. The Adam optimizer is used
with learning rate
$10^{-3}$ and weight decay $10^{-5}$. The 6,000-row chunk size was
selected to fit GPU memory and preserve roughly 10 minutes of
continuous waveform context. The embedding dimensions $d\in\{32,64,128\}$
were evaluated to balance model capacity and computational cost. A \texttt{ReduceLROnPlateau}
scheduler halves the learning rate after 10 epochs without improvement
on the validation metric. Training proceeds for a maximum of 100 epochs
with early stopping (patience~=~10). Mixed-precision (float16) training
via \texttt{torch.cuda.amp} is used throughout.

\section{Experiments and Results}
\label{sec:experiments}


\begin{table*}[t]
\centering
\caption{Stage~1 classification: test-set F1 (label~1) with 95\%
bootstrap CIs for the raw ventilator data}
\label{tab:classification}
\normalsize
\renewcommand{\arraystretch}{1.25}
\setlength{\tabcolsep}{4pt}
\begin{tabular*}{\textwidth}{@{\extracolsep{\fill}}lccccc}
\toprule
\textbf{Configuration} & \textbf{LR} & \textbf{SVM} & \textbf{DT}
    & \textbf{RF} & \textbf{XGB} \\
\midrule
withtime32
    & 0.91 [0.67,1.00]
    & 0.80 [0.33,1.00]
    & 0.75 [0.00,1.00]
    & 0.77 [0.40,1.00]
    & 0.80 [0.40,1.00] \\
withtime64
    & 1.00 [1.00,1.00]
    & 1.00 [1.00,1.00]
    & 1.00 [1.00,1.00]
    & 1.00 [1.00,1.00]
    & 1.00 [1.00,1.00] \\
withtime128
    & 1.00 [1.00,1.00]
    & 1.00 [1.00,1.00]
    & 0.71 [0.36,0.93]
    & 1.00 [1.00,1.00]
    & 1.00 [1.00,1.00] \\
\midrule
notime32
    & 0.57 [0.00,1.00]
    & 0.00 [0.00,0.00]
    & 0.31 [0.00,0.62]
    & 0.20 [0.00,0.53]
    & 0.37 [0.17,0.58] \\
notime64
    & 0.55 [0.00,0.86]
    & 0.00 [0.00,0.00]
    & 0.17 [0.00,0.46]
    & 0.00 [0.00,0.00]
    & 0.00 [0.00,0.00] \\
notime128
    & 0.13 [0.00,0.38]
    & 0.00 [0.00,0.00]
    & 0.11 [0.00,0.32]
    & 0.00 [0.00,0.00]
    & 0.37 [0.17,0.58] \\
\bottomrule
\end{tabular*}
\end{table*}

\begin{table*}[t]
\centering
\caption{Stage~1 classification: test-set F1 (label~1) with 95\%
bootstrap CIs for the jump-point baseline}
\label{tab:jumpbaseline}
\normalsize
\renewcommand{\arraystretch}{1.25}
\setlength{\tabcolsep}{4pt}
\begin{tabular*}{\textwidth}{@{\extracolsep{\fill}}lccccc}
\toprule
\textbf{Configuration} & \textbf{LR} & \textbf{SVM} & \textbf{DT}
    & \textbf{RF} & \textbf{XGB} \\
\midrule
jump\_withtime32  & 0.89 [0.50,1.00] & 0.75 [0.00,1.00] & 0.67 [0.00,1.00] & 0.89 [0.50,1.00] & 0.80 [0.40,1.00] \\
jump\_withtime64  & 0.89 [0.50,1.00] & 0.57 [0.00,1.00] & 0.36 [0.00,0.71] & 0.80 [0.33,1.00] & 0.80 [0.40,1.00] \\
jump\_withtime128 & 0.80 [0.40,1.00] & 0.80 [0.40,1.00] & 0.89 [0.50,1.00] & 0.80 [0.40,1.00] & 0.89 [0.50,1.00] \\
\bottomrule
\end{tabular*}
\end{table*}

\subsection{Experimental Setup}

All experiments use the cohort described in Section~\ref{sec:data}.
We report the two stages of the proposed pipeline separately and then
summarise the final combined configuration. For Stage~1 classification,
the 85-patient cohort is divided into stratified training, validation,
and test sets. Hyperparameters are selected by stratified 5-fold
cross-validation on the combined train\,+\,validation subset, using F1
score on the positive class (label~1) as the primary selection metric.
We compare three embedding dimensions ($d \in \{32,64,128\}$) under two
input settings: \textit{withtime}, in which Time2Vec elapsed-time
encoding is included in the transformer input, and \textit{notime}, in
which that explicit temporal encoding is removed.

For Stage~2 regression, training is restricted to the 29 label~1
patients because only these patients have a defined time-to-event (TTE)
target. The regression branch uses a 19/5/5 train/validation/test
split. Performance is reported on the held-out test set after all model
choices are fixed.

\subsection{Stage 1 Classification Results}

Table~\ref{tab:classification} summarises the Stage~1 classification
results. We observe a clear and consistent pattern: explicit temporal encoding substantially improves classification performance across all model configurations. Across all three embedding dimensions, the
\textit{withtime} models outperform their corresponding \textit{notime}
versions, often by a wide margin. At 32 dimensions, the best
\textit{withtime} model is logistic regression with F1~=~0.91 and a
95\% CI of 0.67,1.00, whereas the best \textit{notime} model at the
same dimension reaches only F1~=~0.57. At 64 and 128 dimensions, several \textit{withtime} models achieve perfect scores on the fixed test split. We interpret this pattern as likely overfitting within the small held-out cohort, and we expect a larger patient sample to provide a more reliable basis for model selection.

For comparison, we evaluate the jump-point baseline from a prior study by members of the current author team~\cite{jumpbaseline2025}, in which raw waveform segments are compressed into 3-dimensional jump-point representations before being passed to a time-aware transformer classifier. Table~\ref{tab:jumpbaseline} shows that the jump-point baseline reaches best F1 values up to 0.89, whereas our best raw-waveform \textit{withtime} Stage~1 models in Table~\ref{tab:classification} perform better on the fixed test split. More importantly, the jump-point representation is inherently limited to classification and cannot directly support the continuous time-to-event estimation required in Stage~2, whereas our approach handles both stages within a unified raw-waveform framework.

To test whether the classifiers behave sensibly away from the final
7-day pre-event window, we apply the same Stage~1 models to the first 7
days of each patient's 30-day recording window, defined as the initial
7 days starting from the first recorded time when the patient used
the home ventilator, and treat those earlier windows as expected
label~0 inputs. Table~\ref{tab:crosswindow} summarises strong
and weak models at each embedding dimension.

The contrast between Tables~\ref{tab:classification} and
\ref{tab:jumpbaseline} also informs our final Stage~1 choice.
Within the 32-dimensional setting, logistic regression
gives the strongest non-degenerate result in Table~\ref{tab:classification},
with F1 = 0.91 and 95\% CI 0.67,1.00. XGBoost is the
next-strongest 32-dimensional \textit{withtime} alternative at F1 = 0.80
and remains stable in the first-7-day check. Although logistic regression gives
the best positive-class F1 on the final-window test set, its weak behavior on the
early-window stability check motivates retaining XGBoost as a complementary
secondary classifier. We therefore use the
32-dimensional \textit{withtime} setting with logistic regression as the
primary Stage~1 classifier and XGBoost as a secondary classifier for stability verification.

\begin{table}[t]
\centering
\caption{Cross-window stability check on the first 7 days of each patient's 30-day recording window}
\label{tab:crosswindow}
\normalsize
\setlength{\tabcolsep}{4pt}
\renewcommand{\arraystretch}{1.25}
\resizebox{\columnwidth}{!}{%
\begin{tabular}{lccccc}
\toprule
\textbf{Dim} & \textbf{Setting} & \textbf{Model} & \textbf{AccL0} & \textbf{Label~0} & \textbf{Label~1} \\
\midrule
32  & Strong & XGBoost & 0.97 & 78 & 7 \\
32  & Weak   & Logistic Regression & 0.00 & 0 & 85 \\
32  & Strong & Decision Tree & 0.97 & 83 & 2 \\
\midrule
64  & Strong & XGBoost & 1.00 & 85 & 0 \\
64  & Strong & Logistic Regression & 1.00 & 85 & 0 \\
64  & Weak   & SVM & 0.00 & 0 & 85 \\
\midrule
128 & Strong & SVM & 1.00 & 85 & 0 \\
128 & Strong & Random Forest & 0.95 & 76 & 9 \\
128 & Weak   & Decision Tree & 0.00 & 0 & 85 \\
\bottomrule
\end{tabular}%
}
\end{table}

\subsection{Stage 2 Regression Results}

\begin{table}[t]
\centering
\caption{Stage~2 regression: test-set RMSE(Days), MAE(Days), and R$^2$ with 95\% bootstrap CIs for the raw ventilator data}
\label{tab:regression}
\normalsize
\renewcommand{\arraystretch}{1.25}
\setlength{\tabcolsep}{4pt}
\makebox[\columnwidth][c]{%
\resizebox{1.00\columnwidth}{!}{%
\begin{tabular}{lccc}
\toprule
\textbf{Model}
    & \textbf{RMSE}
    & \textbf{MAE}
    & \textbf{R$^2$} \\
\midrule
\multicolumn{4}{l}{\textit{Time-Aware Transformer}} \\
32dim
    & 1.16 [0.97,1.34]
    & 1.04 [0.86,1.21]
    & 0.68 [0.51,0.76] \\
64dim
    & 1.00 [0.85,1.14]
    & 0.87 [0.70,1.03]
    & 0.76 [0.62,0.84] \\
128dim
    & 1.04 [0.87,1.19]
    & 0.91 [0.74,1.07]
    & 0.74 [0.58,0.83] \\
\midrule
\multicolumn{4}{l}{\textit{Baselines (all embedding dimensions)}} \\
Mean
    & 2.05 [1.75,2.34]
    & 1.80 [1.45,2.13]
    & 0.00 [$-$0.16,0.00] \\
Median
    & 2.05 [1.74,2.33]
    & 1.79 [1.42,2.15]
    & 0.00 [$-$0.16,0.00] \\
Ridge
    & 2.25 [1.87,2.59]
    & 1.96 [1.57,2.33]
    & $-$0.20 [$-$0.63,$-$0.01] \\
XGBoost
    & 2.12 [1.65,2.55]
    & 1.76 [1.37,2.17]
    & $-$0.06 [$-$0.49,0.21] \\
\bottomrule
\end{tabular}
}}
\end{table}

 Table\ref{tab:regression} reports the Stage~2 regression results on the label~1 patients. These test-set results are derived from the leave-one-patient-out cross-validation (LOPO-CV) style evaluation. We first compare the three
Time-Aware Transformer regression models and then report
non-transformer baselines built from per-day waveform statistics. Among
the transformer settings, we select the 64-dimensional model because it
achieves the lowest RMSE and MAE and the highest $R^2$ with bootstrap
confidence intervals reported in the table. All transformer variants
also outperform the summary-statistic baselines, whose $R^2$ values are
near zero or negative. The mean and median baselines produce RMSE values near
2.05 days and $R^2$ values close to zero, indicating that simple central-tendency
predictors explain little patient-specific TTE variation. Ridge regression performs
worse, with RMSE = 2.25 days and negative $R^2$, suggesting that the summary
features do not support a reliable linear mapping to TTE in this cohort. XGBoost
is the strongest non-transformer baseline but still remains well below the
64-dimensional transformer, which supports using the raw-waveform temporal model
rather than only aggregated waveform statistics.

Based on these results, we use the 64-dimensional model as
our Stage~2 regression model. It achieves RMSE = 1.00 days
(95\% CI 0.85,1.14), MAE = 0.87 days (95\% CI
0.70,1.03), and $R^2$ = 0.76 (95\% CI 0.62,0.84).
The 128-dimensional model remains close but does not improve on the
64-dimensional setting.

To translate the regression output into a clinically usable alert, we
evaluate thresholds from 1 to 5 days on all 29 label~1 patients.
Table~\ref{tab:threshold} suggests that a threshold of $<3$ days provides
the most balanced trade-off among early warning sensitivity, alert
precision, and specificity in this cohort. At this threshold, sensitivity is 0.96,
specificity is 0.70, PPV is 0.72, and NPV is 0.96. Stricter
thresholds reduce PPV, whereas more relaxed thresholds increase
sensitivity at the cost of a much heavier false-alert burden.

\begin{table}[t]
\centering
\caption{Stage~2 alert threshold: sensitivity, specificity, PPV, and NPV on the 29 label-1 patients (embed\_dim=64)}
\label{tab:threshold}
\normalsize
\renewcommand{\arraystretch}{1.2}
\setlength{\tabcolsep}{5pt}
\begin{tabular}{ccccc}
\toprule
\textbf{Threshold} & \textbf{Sens.} & \textbf{Spec.} & \textbf{PPV} & \textbf{NPV} \\
\midrule
$<1$ day            & 0.97 & 0.78 & 0.44 & 0.99 \\
$<2$ days           & 0.96 & 0.75 & 0.62 & 0.98 \\
$<3$ days           & 0.96 & 0.70 & 0.72 & 0.96 \\
$<4$ days           & 0.99 & 0.24 & 0.63 & 0.95 \\
$<5$ days           & 1.00 & 0.00 & 0.72 & ---  \\
\bottomrule
\end{tabular}
\end{table}

\begin{figure}[t]
    \centering
    \includegraphics[width=0.88\columnwidth]{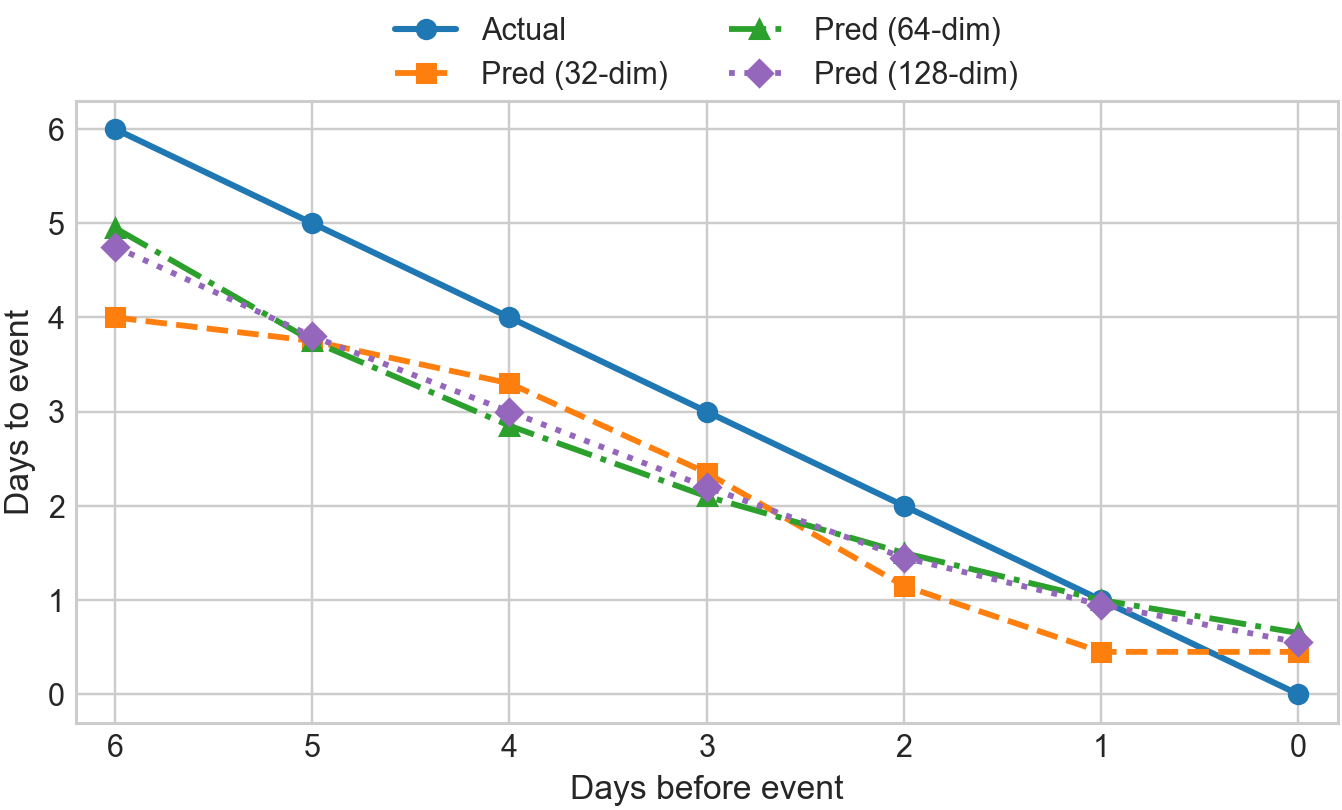}
    \caption{Illustrative time-to-event prediction example for one
    representative test patient, not an averaged trajectory. The same
    patient is evaluated with the 32-, 64-, and 128-dimensional
    time-aware regression models. The x-axis is shown in reverse order
    from 6 to 0 days before event.}
    \label{fig:tte_examples}
\end{figure}

This threshold analysis is consistent with the patient-level example in
Figure~\ref{fig:tte_examples}. This figure is shown as an
illustrative prediction for one label~1 patient. For the illustrated test patient,
the prediction error narrows as the event approaches, with the most accurate
estimates in the final 3 days before exacerbation.

\subsection{Combined Pipeline Performance}

The final two-stage configuration combines what we view as the most defensible Stage~1
setting with the strongest Stage~2 regression model. Specifically,
Stage~1 uses the 32-dimensional withtime transformer
embedding, with logistic regression as the primary classifier and
XGBoost as a secondary classifier for stability verification, and Stage~2 uses the
64-dimensional time-aware regression model. In deployment, a
patient is first assigned to label~0 or label~1 from the last 7 days of
raw pressure and flow waveforms. Patients predicted as
label~0 receive no alert. Patients predicted as label~1 are passed to
Stage~2, which outputs the estimated remaining time to event and raises
an alert when predicted TTE is below 3 days.

This final combination reflects the different requirements of the two
stages. For Stage~1, we prioritise a conservative and interpretable
operating point rather than the most optimistic fixed-split score,
which leads us to the 32-dimensional logistic regression model while
retaining XGBoost as a secondary stability check. For
Stage~2, the 64-dimensional transformer gives the best held-out TTE
accuracy and supports a practical 3-day alert threshold. Overall, we believe the
pipeline suggests that 7 days of raw home-ventilator waveforms can support
both high-risk screening and short-horizon event-timing estimation in a
single clinically oriented framework.
\section{Discussion}
\label{sec:discussion}

\subsection{Clinical Implications}

We design this framework to transform raw home ventilator data into two
clinically actionable outputs: (1) a binary risk flag indicating whether
a patient is trending towards a severe event, and (2) a time-to-event
estimate of days remaining before the predicted event. In our experiments, the average lead time is approximately three days. We believe this window allows
clinicians to intensify therapy, arrange a clinic visit, or prepare for
potential hospital transfer. Such actions have been shown to reduce
exacerbation severity and length of stay \cite{rueda2024machine}.

A key practical contribution of this study is the design of the
two-stage model, which first applies
column selection to raw \texttt{pressure} and \texttt{flow}, then
performs classification, and finally time-to-event regression. Low latency means that the framework relies on
continuously available home ventilator waveforms instead of delayed
clinical or laboratory measurements.

This pressure--flow focus is clinically motivated. Pressure and flow are
the two ventilator waveforms most routinely inspected at the bedside,
and they directly reflect respiratory mechanics, patient effort,
triggering/cycling behavior, and air trapping
\cite{hess2005waveforms,fernandez2006pressure,hamahata2020flow}. In
COPD specifically, flow-based measures such as peak expiratory flow have
also been associated with exacerbation detection and hospitalization
assessment \cite{cen2019pef,cen2022pef}. Transformer inference over a
6,000-row chunk requires only a few seconds on GPU. From a Human-Machine Systems perspective, our two-stage model can reduce reliance on delayed laboratory measurements and serve as a decision-support tool for prioritizing high-risk patients, while leaving final clinical judgment to the clinicians.

\subsection{Limitations}

Several limitations must be acknowledged. The cohort contains 87 patients
with an approximately 2:1 class imbalance, which is small by deep learning
standards. Hyperparameter optimization was performed with 5-fold
cross-validation on the combined training and validation set to reduce
overfitting risk, but the 22-patient test set still provides limited
statistical power.
Baseline demographic characteristics were not available because of privacy restrictions.

The current model uses only two input columns, \texttt{flow} and
\texttt{pressure}. This restriction is intentional rather than arbitrary,
because these are the primary ventilator scalars used in routine waveform
interpretation and capture much of the information most relevant to
obstruction, resistance, and patient--ventilator interaction
\cite{hess2005waveforms,fernandez2006pressure,hamahata2020flow}. Using
only two channels also improves GPU efficiency and keeps training and
inference computationally manageable on long raw waveform sequences.

External validation is also challenging in this setting. As illustrated by
the waveform example in Figure~\ref{fig:waveform_example}, our ventilator
recordings are sampled at 5 readings per second, which preserves
within-breath shape and rapid cycle-to-cycle variation in \texttt{flow}
and \texttt{pressure}. Many public datasets
either use very different devices, different waveform definitions, or much
lower and incompatible sampling schemes.
Hospitals and device manufacturers should standardize ventilator waveform formats and collect higher-frequency data to support more reliable external validation.

Finally, due to the structure of the time-aware transformers, model interpretability remains limited, and its predictions
should be used only as decision support while clinicians make the final
clinical decisions.

\section{Conclusion}
\label{sec:conclusion}

This paper presented the Two-Stage AECOPD Time-Aware Transformer Model
for prediction from home ventilator data over a 7-day pipeline. The
final system uses raw \texttt{pressure} and \texttt{flow} waveforms as
input. Our selected configuration combines a 32-dimensional time-aware
embedding with logistic regression as the primary Stage~1 classifier
and XGBoost as a secondary classifier for stability verification, together with a
64-dimensional time-aware transformer regression model for Stage~2
time-to-event estimation. On the held-out test set, the selected
Stage~1 classifier achieves F1 = 0.91 for the high-risk group
(label~1), and the selected Stage~2 regression model achieves
RMSE = 1.00 days, MAE = 0.87 days, and $R^2$ = 0.76.

In our view, these results suggest that the framework can distinguish
high-risk patients and provide short-horizon timing information once a
patient is flagged as high risk. Compared with the jump-point baseline,
we find that the raw-waveform approach gives stronger classification
performance while preserving the temporal continuity required for
regression. Overall, we believe the model supports clinically actionable
home monitoring over the most recent 7-day window, although larger
cohorts are still needed to confirm its generalizability. By combining
two-stage prediction with continuous home ventilator monitoring, this
framework may support more efficient early identification and intervention
for telemedicine and aging-in-place care.

\bibliographystyle{IEEEtran}
\bibliography{main}

@article{rueda2024machine,
  author  = {Rueda, R. and Fabello, E. and Silva, T. and Genzor, S. and Mizera, J. and Stanke, L.},
  title   = {Machine learning approach to flare-up detection and clustering in chronic obstructive pulmonary disease ({COPD}) patients},
  journal = {Health Information Science and Systems},
  year    = {2024},
  volume  = {12},
  number  = {1},
  pages   = {50},
  doi     = {10.1007/S13755-024-00308-4}
}

@article{lenoir2023mortality,
  author  = {Lenoir, A. and Whittaker, H. and Gayle, A. and Jarvis, D. and Quint, J. K.},
  title   = {Mortality in non-exacerbating {COPD}: a longitudinal analysis of {UK} primary care data},
  journal = {Thorax},
  year    = {2023},
  volume  = {78},
  number  = {9},
  pages   = {904--911},
  doi     = {10.1136/thorax-2022-218724}
}

@article{kor2022explainable,
  author  = {Kor, C. T. and Li, Y. R. and Lin, P. R. and Lin, S. H. and Wang, B. Y. and Lin, C. H.},
  title   = {Explainable machine learning model for predicting first-time acute exacerbation in patients with chronic obstructive pulmonary disease},
  journal = {Journal of Personalized Medicine},
  year    = {2022},
  month   = feb,
  volume  = {12},
  number  = {2},
  pages   = {228},
  doi     = {10.3390/jpm12020228}
}

@article{hess2005waveforms,
  author  = {Hess, Dean R.},
  title   = {Ventilator waveforms and the physiology of pressure support ventilation},
  journal = {Respiratory Care},
  year    = {2005},
  month   = feb,
  volume  = {50},
  number  = {2},
  pages   = {166--186},
  doi     = {10.4187/respcare.05500166}
}

@article{fernandez2006pressure,
  author  = {Fernandez-P{\'e}rez, Evans R. and Hubmayr, Rolf D.},
  title   = {Interpretation of airway pressure waveforms},
  journal = {Intensive Care Medicine},
  year    = {2006},
  month   = may,
  volume  = {32},
  number  = {5},
  pages   = {658--659},
  doi     = {10.1007/s00134-006-0108-7}
}

@article{hamahata2020flow,
  author  = {Hamahata, Natsumi T. and Sato, Ryota and Daoud, Ehab G.},
  title   = {Go with the flow---clinical importance of flow curves during mechanical ventilation: a narrative review},
  journal = {Canadian Journal of Respiratory Therapy},
  year    = {2020},
  month   = jul,
  volume  = {56},
  pages   = {11--20},
  doi     = {10.29390/cjrt-2020-002}
}

@article{cen2019pef,
  author  = {Cen, Jie and Ma, Hongying and Chen, Zhongbo and Weng, Lei and Deng, Zaichun},
  title   = {Monitoring peak expiratory flow could predict {COPD} exacerbations: a prospective observational study},
  journal = {Respiratory Medicine},
  year    = {2019},
  month   = mar,
  volume  = {148},
  pages   = {43--48},
  doi     = {10.1016/j.rmed.2019.01.010}
}

@article{cen2022pef,
  author  = {Cen, Jie and Weng, Lei},
  title   = {Comparison of peak expiratory flow ({PEF}) and {COPD} assessment test ({CAT}) to assess {COPD} exacerbation requiring hospitalization: a prospective observational study},
  journal = {Chronic Respiratory Disease},
  year    = {2022},
  month   = feb,
  volume  = {19},
  pages   = {14799731221081859},
  doi     = {10.1177/14799731221081859}
}

@article{jiang2021nppv,
  author  = {Jiang, Weipeng and Chao, Yencheng and Wang, Xiaoyue and Chen, Cuicui and Zhou, Jian and Song, Yuanlin},
  title   = {Day-to-day variability of parameters recorded by home noninvasive positive pressure ventilation for detection of severe acute exacerbations in {COPD}},
  journal = {International Journal of Chronic Obstructive Pulmonary Disease},
  year    = {2021},
  month   = mar,
  volume  = {16},
  pages   = {727--737},
  doi     = {10.2147/COPD.S299819}
}

@article{wu2021upcoming7days,
  author   = {Wu, Chia-Tung and Li, Guo-Hung and Huang, Chun-Ta and Cheng, Yu-Chieh and Chen, Chi-Hsien and Chien, Jung-Yien and Kuo, Ping-Hung and Kuo, Lu-Cheng and Lai, Feipei},
  title    = {Acute Exacerbation of a Chronic Obstructive Pulmonary Disease Prediction System Using Wearable Device Data, Machine Learning, and Deep Learning: Development and Cohort Study},
  journal  = {JMIR mHealth and uHealth},
  year     = {2021},
  month    = may,
  volume   = {9},
  number   = {5},
  doi      = {10.2196/22591}
}

@article{kazemi2019time2vec,
  author  = {Kazemi, S. M. and Goel, R. and Eghbali, S. and Ramanan, J. and Sahota, J. and Thakur, S. and others},
  title   = {{Time2Vec}: Learning a vector representation of time},
  journal = {arXiv preprint arXiv:1907.05321},
  year    = {2019},
  month   = jul,
  doi     = {10.48550/arXiv.1907.05321}
}

@article{foumani2024tape,
  author  = {Foumani, Navid Mohammadi and Tan, Chang Wei and Webb, Geoffrey I. and Salehi, Mahsa},
  title   = {Improving position encoding of transformers for multivariate time series classification},
  journal = {Data Mining and Knowledge Discovery},
  year    = {2024},
  month   = jan,
  volume  = {38},
  pages   = {22--48},
  doi     = {10.1007/s10618-023-00948-2}
}

@article{jumpbaseline2025,
  author  = {Qu, W. and Zheng, L. and Wang, D. and Wang, J. and Pan, H.},
  title   = {Time-aware transformer-based prediction model for {AECOPD}},
  journal = {Studies in Health Technology and Informatics},
  year    = {2025},
  volume  = {329},
  pages   = {1089--1093},
  doi     = {10.3233/SHTI251007}
}

@inproceedings{zeng2023transformers,
  author    = {Zeng, Ailing and Chen, Muxi and Zhang, Lei and Xu, Qiang},
  title     = {Are Transformers Effective for Time Series Forecasting?},
  booktitle = {AAAI Conference on Artificial Intelligence},
  year      = {2023},
  volume    = {37},
  number    = {9},
  pages     = {11121--11128},
  doi       = {10.1609/aaai.v37i9.26317}
}

@inproceedings{ICLR2025_2b187165,
  author    = {Wang, Shiyu and Li, Jiawei and Shi, Xiaoming and Ye, Zhou and Mo, Baichuan and Lin, Wenze and Ju, Shengtong and Chu, Zhixuan and Jin, Ming},
  title     = {{TimeMixer++}: A general time series pattern machine for universal predictive analysis},
  booktitle={International Conference on Learning Representations},
  volume={2025},
  pages={16980--17016},
  year={2025}
}

@article{yin2024inhome,
  author  = {Yin, Huiming and Wang, Kun and Yang, Ruyu and Tan, Yanfang and Li, Qiang and Zhu, Wei and Sung, Suzi},
  title   = {A machine learning model for predicting acute exacerbation of in-home chronic obstructive pulmonary disease patients},
  journal = {Computer Methods and Programs in Biomedicine},
  year    = {2024},
  volume  = {246},
  pages   = {108005},
  doi     = {10.1016/j.cmpb.2023.108005}
}

@article{liao2024nearfuture,
  author  = {Liao, Kuang-Ming and Cheng, Kuo-Chen and Sung, Mei-I and Shen, Yu-Ting and Chiu, Chong-Chi and Liu, Chung-Feng and Ko, Shian-Chin},
  title   = {Machine learning approaches for practical predicting outpatient near-future {AECOPD} based on nationwide electronic medical records},
  journal = {iScience},
  year    = {2024},
  volume  = {27},
  number  = {4},
  pages   = {109542},
  doi     = {10.1016/j.isci.2024.109542}
}

@article{tyrovolas2026efficient,
  author  = {Tyrovolas, Marios and Kallimanis, Nikolaos D. and Stylios, Chrysostomos},
  title   = {Efficient Total Causal Effect Computation in Fuzzy Cognitive Maps for Scalable Explainable Artificial Intelligence},
  journal = {IEEE Systems, Man, and Cybernetics Letters},
  year    = {2026},
  doi     = {10.1109/LSMC.2026.3692435}
}

@article{li2026behavior,
  author  = {Li, Yuan and Zhou, Yuhang and Dai, Shibo and Wang, Jianyu and Wu, Xiang},
  title   = {Behavior-Guided Identity Learning for Multiagent Cooperation},
  journal = {IEEE Systems, Man, and Cybernetics Letters},
  year    = {2026},
  doi     = {10.1109/LSMC.2026.3696310}
}

@article{qu2026multimodal,
  author  = {Qu, Weihao and Wang, Dongyang and Zheng, Ling and Alvarez, Francisco E. and Polasa, Shobharani and Wang, Jiacun},
  title   = {Multimodal Injury Risk and Performance Prediction in Tennis Using Weighted Ensemble Learning},
  journal = {IEEE Systems, Man, and Cybernetics Magazine},
  year    = {2026},
  pages   = {1--7},
  doi     = {10.1109/MSMC.2026.3685426},
  note    = {Early Access}
}

\end{document}